\documentclass[runningheads]{llncs}

\usepackage{eccv}

\usepackage{eccvabbrv}

\usepackage{graphicx}
\usepackage{booktabs}
\usepackage{multirow}
\usepackage{colortbl}

\newcommand{\bestCellColor}[1]{\cellcolor[rgb]{.866,.945, 0.831}#1}

\newcommand{\secondBestCellColor}[1]{\cellcolor[rgb]{1, 0.98, 0.83}#1}

\newcommand{\abbtitle}{Director\xspace}

\usepackage{soul}
\definecolor{junglegreen}{rgb}{0.113, 0.639, 0.5}

\usepackage[accsupp]{axessibility}  %

\usepackage{hyperref}

\usepackage{orcidlink}

\definecolor{cvprblue}{rgb}{0.21,0.49,0.74}
\definecolor{custompink}{RGB}{255, 0, 144}  %

\newcommand{\myparagraph}[1]{\vspace{0.1em}\noindent\textbf{#1}}

\begin{document}

\title{\abbtitle: Instance-aware Gaussian Splatting for Dynamic Scene Modeling and Understanding} 

\titlerunning{\abbtitle}

\author{Yuheng Jiang\inst{1}\textsuperscript{*} \and  Yiwen Cai\inst{1}\textsuperscript{*} \and Zihao Wang\inst{2} \and Yize Wu\inst{1} \and Sicheng Li\inst{2} \and Zhuo Su\inst{2} \and  Shaohui Jiao\inst{2} \and Lan Xu\inst{1}}

\authorrunning{Y. Jiang et al.}

\institute{ShanghaiTech University \and ByteDance}

\maketitle

\begin{center}
    \vspace{-3ex}
    \includegraphics[width=0.97\textwidth]{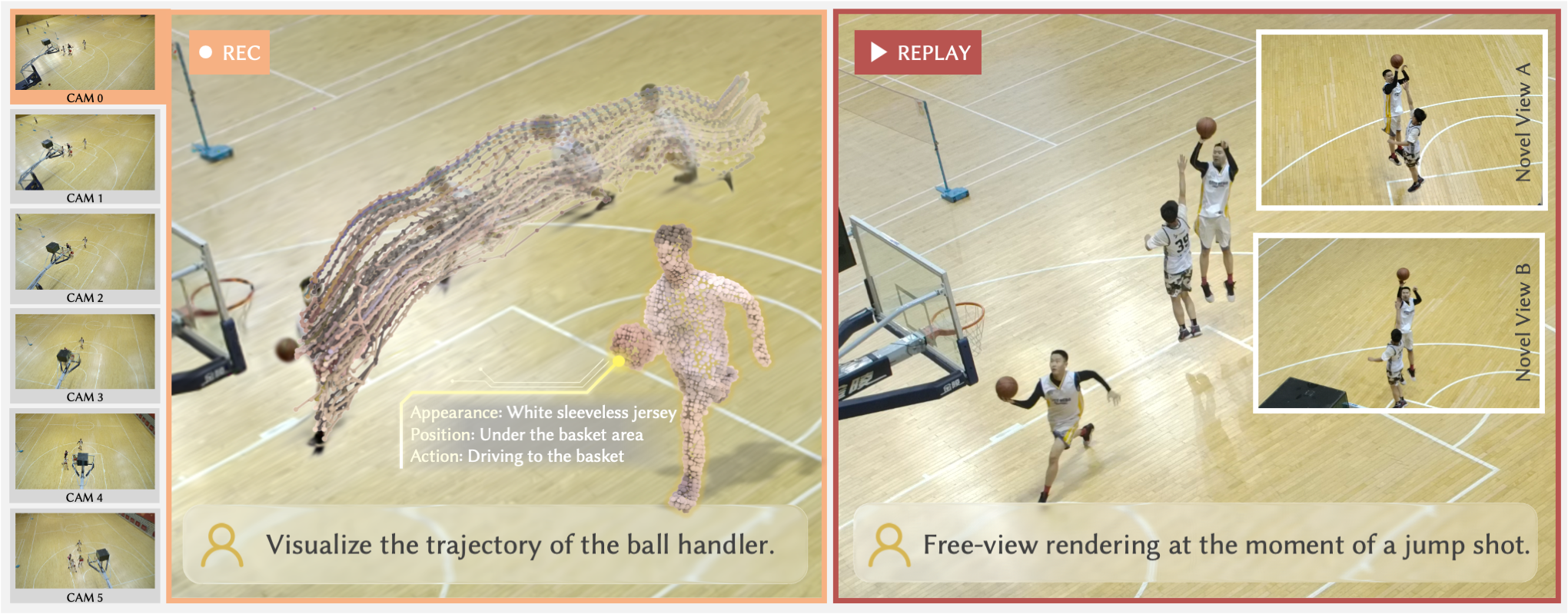}
    \captionof{figure}{We introduce \textbf{\abbtitle}, an instance-aware spatio-temporal Gaussian representation that enables robust human performance tracking, high-fidelity rendering, and instance-level understanding for open-vocabulary queries.}
    \label{fig:teaser}
\end{center}

\begin{abstract}
  Volumetric video seeks to model dynamic scenes as temporally coherent 4D representations. While recent Gaussian-based approaches achieve impressive rendering fidelity, they primarily emphasize appearance but are largely agnostic to instance-level structure, limiting stable tracking and semantic reasoning in highly dynamic scenarios.
In this paper, we present \abbtitle, a unified spatio-temporal Gaussian representation that jointly models human performance, high-fidelity rendering, and instance-level semantics. 
Our key insight is that embedding instance-consistent semantics naturally complements 4D modeling, enabling more accurate scene decomposition while supporting robust dynamic scene understanding.
To this end, we leverage temporally aligned instance masks and sentence embeddings derived from Multimodal Large Language Models to supervise the learnable semantic features of each Gaussian via two MLP decoders, enabling language-aligned 4D representations and enforcing identity consistency over time.
To enhance temporal stability, we bridge 2D optical flow with 4D Gaussians and finetune their motions, yielding reliable initialization and reducing drift.
For the training, we further introduce a geometry-aware SDF constraints, along with regularization terms that enforces surface continuity, enhancing temporal coherence in dynamic foreground modeling.
Experiments demonstrate that \abbtitle achieves temporally coherent 4D reconstructions while simultaneously enabling instance segmentation and open-vocabulary querying.
Project page: \href{https://caiyw2023.github.io/Director/}{\textcolor{custompink}{https://caiyw2023.github.io/Director/.}}

\end{abstract}

\section{Introduction}

We see what moves. We shape what persists. We understand what matters.
Volumetric video aspires to capture this very progression, transforming dynamic visual observations into structured 4D representations, laying the foundation for scene understanding.

Yet, existing solutions~\cite{collet2015high, isik2023humanrf, sun20243dgstream, gao2024gaussianflow} largely optimize dynamic scenes for photometric consistency, evolving from mesh-based models to NeRF~\cite{nerf} and Gaussian Splatting~\cite{3dgs}. Despite achieving remarkable rendering fidelity, these approaches remain largely agnostic to the underlying scene structure, treating scenes as collection of radiance primitives without explicit instance-consistent constraints. Consequently, in highly dynamic scenarios, the optimization process often suffers from identity drift, foreground–background entanglement, unstable semantic association over time.
A few recent works~\cite{sa4d, li2024sadg} begin to move toward deeper structural understanding by introducing instance-level segmentation. Another line of research~\cite{kerr2023lerf, li20254d} explores natural language as an interface for scene querying and interaction. However, these approaches still adhere to a “reconstruction-then-understanding” paradigm, treating geometric modeling and semantics as loosely coupled components rather than mutually reinforcing processes. Moreover, they typically handle only carefully curated slow-motion scenarios and remain fragile when confronted with complex or fast dynamic motions.

In this paper, we argue that dynamic reconstruction and understanding should not follow the reconstruction-then-understanding pipeline, but instead be jointly optimized through instance-consistent primitive-level modeling.
Instance-level segmentation and semantic cues enable more precise dynamic–static or inter-instance decomposition, improving tracking and appearance modeling. Conversely, long-term tracking and coherent 4D modeling provide temporally consistent context that enhances segmentation quality and forms a reliable foundation for open-vocabulary queries over dynamic scenes.

To ``shape what persists and understand what matters'', we introduce a unified and instance-aware spatio-temporal Gaussian representation for highly dynamic scenes~(see Fig.~\ref{fig:teaser}). Our core idea is to leverage a set of temporally consistent 4D Gaussian primitives equipped with learnable semantic feature attributes to represent the scene. Guided by SAM3-derived instance masks and MLLM-derived semantic embedding, the Gaussian primitives are optimized to achieve accurate tracking, high-fidelity rendering, and detailed instance-level segmentation.

Our method, \abbtitle, first establishes reliable instance correspondences across multiple camera views. Leveraging temporally consistent SAM3~\cite{sam3} masks, we resolve cross-view ambiguities to obtain spatially and temporally consistent instance masks. To “understand what matters”, we generate time-varying, instance-wise captions from Multimodal Large Language Models (MLLMs) and encode them into sentence embeddings, which serve as language- and instance-aligned supervision for the training of the dynamic semantic field.

With the instance masks and semantic embedding as guidance, we explicitly decompose the scene into a static background and a dynamic foreground to enhance the consistency.
For the static background, we aggregate observations across the entire temporal sequence and optimize a temporally consistent Gaussian layer. This provides a global reference for the scene, and facilitates accurate separation of dynamic foreground elements.
For the dynamic foreground, we equip each Gaussian with a learnable semantic feature vector and use two MLP decoders to predict instance identity and language-aligned semantics. To enforce spatial coherence, we apply a KL-divergence regularization that encourages neighboring Gaussians within the same instance to maintain consistent semantic features.

To handle fast and complex motions in highly dynamic scenes, we explicitly warp the Gaussians from the previous frame using optical flow, providing reliable position initialization.
During training, we first fine-tune only the Gaussian positions and rotations with an as-rigid-as-possible regularizer, enforcing local geometric consistency and correcting misalignments. 
We then leverage geometry-aware SDF constraints and regularization terms to jointly optimize Gaussian motion, appearance, and semantic features, maintaining instance-level consistency and temporal coherence.
After training, our method identifies the target instance via identity query, renders its instance-level feature map, and retrieves corresponding relevant frame segments with segment queries, effectively filtering out interference from other dynamic objects.

To summarize, our main contributions include:
\begin{itemize} 
	\setlength\itemsep{0em}
	
	\item 
    
    A primitive-level instance-consistent 4D Gaussian formulation that tightly couples dynamic reconstruction and semantic modeling within a unified representation.

	\item 
    
    Language- and instance-aligned semantic features embedded into dynamic Gaussians, where instance-level mask supervision and language priors jointly enforce structural constraints for temporally coherent 4D modeling, enabling detailed instance segmentation and open-vocabulary querying.

	\item A two-layer Gaussian representation that explicitly separates static and dynamic components and incorporates optical-flow-guided optimization for robust human performance tracking and high-fidelity rendering.

\end{itemize}

\section{Related Work} 

\noindent{\textbf{Non-Rigid Tracking.}} 
Recent research on human performance non-rigid tracking has been extensively explored for various applications~\cite{xiang2020monoclothcap, LiveCap2019tog, Wang2021CVPR, li2021deep, zhang2023closet, zhao2022human, shao2022floren, guo2017real, xu2019deep, realTimeDDC, kwon2024deliffas, sun2021HOI-FVV, suo2021neuralhumanfvv, jiang2022neuralhofusion}.
DynamicFusion~\cite{newcombe2015dynamicfusion} pioneered this direction by mapping dynamic frames into a canonical space via a dense volumetric warp field. VolumeDeform~\cite{innmann2016volume} enhanced robustness via sparse color feature matching. Fusion4D~\cite{dou2016fusion4d} scaled to multi-view capture for large motions and topological changes. KillingFusion and SobolevFusion~\cite{KillingFusion2017cvpr, slavcheva2018sobolevfusion} replaced explicit correspondence estimation with differential regularization on signed distance fields, and DoubleFusion~\cite{DoubleFusion} embedded a parametric body model into a dual-layer representation for better occlusion handling. Later, Motion2Fusion~\cite{motion2fusion} and UnstructureFusion~\cite{UnstructureLan} accelerated capture through learned correspondences in monocular and unstructured multi-camera settings, while RobustFusion~\cite{robustfusion, su2022robustfusionPlus} addressed human-object interactions via semantics-aware scene decomposition. DDC~\cite{realTimeDDC} further shifted toward weakly supervised learning, achieving photorealistic real-time reconstruction with fine-grained clothing details. Dynamic 3D Gaussians~\cite{luiten2023dynamic} allowed Gaussians to move and rotate over time to represent scene motions. 
Notably, recent works~\cite{wang2023omnimotion, jin2025stereo4d, li2025megasam, st4rtrack2025, cut3r} have made significant progress in point tracking from casual monocular videos. Building on this line of work, our method incorporates multi-view 2D semantic priors and optical flow to maintain robust tracking in highly dynamic scenes with complex interactions and challenging occlusions.

\noindent{\textbf{Dynamic Scene Representation.}} 
Dynamic scene representation has undergone a fundamental paradigm shift over the past decade. Early mesh-based methods~\cite{collet2015high, dou2016fusion4d, motion2fusion, yu2021function4d} reconstructed textured dynamic meshes from dense camera rigs, but struggled to capture view-dependent appearance. Neural Radiance Fields (NeRF)~\cite{nerf} substantially elevated rendering fidelity through implicit volumetric representations, with dynamic extensions following either canonical-space deformation~\cite{park2021nerfies, park2021hypernerf, li2021neural, zhao2022human, luo2022artemis} or spatio-temporal tensor factorization~\cite{cao2023hexplane, fridovich2023k, shao2023tensor4d, isik2023humanrf, song2023nerfplayer}. Despite their quality gains, NeRF-based methods remain bottlenecked by costly ray-marching, limiting practical efficiency. 3D Gaussian Splatting (3DGS)~\cite{3dgs} has since enabled photorealistic quality with fast training and real-time rendering, spurring three main directions for dynamic scenes: (1) per-Gaussian motion tracking across frames~\cite{luiten2023dynamic, jiang2024hifi4g, dualgs, sun20243dgstream,taogs, wang2025shape, rai2026packuv, hong2025beam}, (2) canonical Gaussian representations with deformation fields~\cite{yang2024deformable, Wu_2024_CVPR, jiang2025reperformer, qian20243dgs, pang2024ash, li2025gifstream}, and (3) native 4D spacetime Gaussian representations~\cite{yang2023real, duan20244d, li2023spacetime}, which parameterize Gaussians directly in 4D for real-time rendering via anisotropic temporal slicing.
Our work follows the per-Gaussian motion tracking paradigm, augmenting it with semantic knowledge distilled from imperfect 2D semantic masks to achieve robust, high-quality dynamic scene representation.

\noindent{\textbf{Instance-level Scene Understanding and Editing.}} 
Bridging scene representations with 2D vision–language foundation models~\cite{hurst2024gpt, bai2025qwen3, wang2024qwen2} through language fields has emerged as a dominant paradigm for open-vocabulary 3D scene understanding. Early works~\cite{peng2023openscene, kobayashi2022decomposing, vora2021nesf,wang2022clip, yu2021unsupervised, kania2022conerf, yuan2022nerf} lift 2D features into static 3D scenes. LERF~\cite{kerr2023lerf} distills CLIP features into NeRF to enable language-driven queries. With the advent of 3D Gaussian Splatting (3DGS)~\cite{3dgs}, LangSplat~\cite{qin2024langsplat} integrates 3DGS with SAM-derived hierarchical semantic maps for more precise language field modeling.
Recent works~\cite{sa4d} extend this paradigm to dynamic settings. 4DLangSplat~\cite{li20254d} supervises per-object trajectories using temporally consistent captions and models state transitions via deformation networks. 4-LEGS~\cite{fiebelman20254legs} embeds spatio-temporal video features into 4D Gaussians to support text-based event localization across space and time. 4DLangVGGT~\cite{wu20254dlangvggt} builds upon VGGT~\cite{wang2025vggt}, a feed-forward visual geometry foundation model, and jointly learns spatio-temporal geometry and language alignment, enabling efficient open-vocabulary 4D understanding without per-scene optimization.
Inspired by these advances, our work leverages long-term tracking and coherent 4D modeling to enhance segmentation quality and establish a reliable foundation for open-vocabulary queries over dynamic scenes.

\newpage

\begin{figure*}[ht] 
	\begin{center} 
		\includegraphics[width=\linewidth]{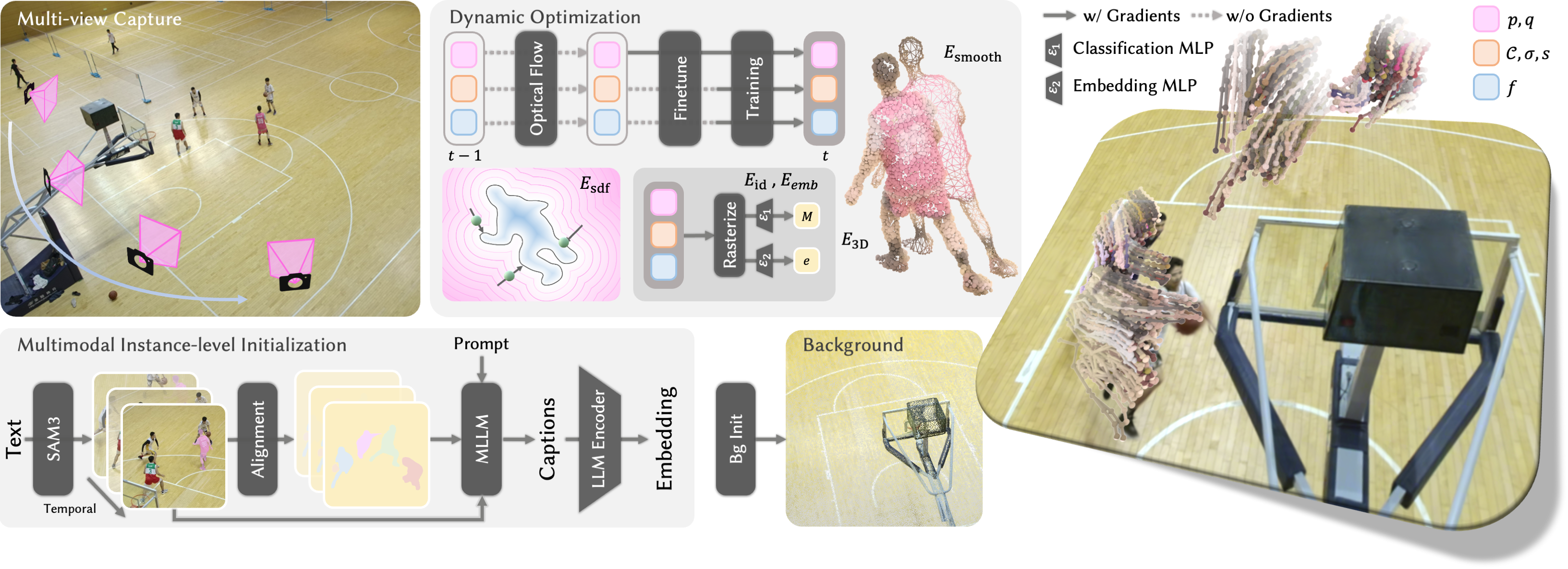} 
	\end{center} 
    \vspace{-20pt}  
	\caption{\noindent{\bf Overview of \abbtitle.} Using temporally consistent SAM3 masks and sentence embeddings, our method decomposes the scene into static background and dynamic foreground, learning language- and instance-aligned features for robust tracking, high-quality rendering, and accurate instance segmentation. }
	\label{fig:overview} 
	\vspace{-35pt}
\end{figure*}

\section{Method}\label{sec:algorithm} 
Our method is built upon a key design principle:
instance-level consistency should be enforced directly at the Gaussian primitive level.
To achieve this, we embed instance identities and language-aligned sentences embeddings into each Gaussian’s semantic feature,
and jointly optimize them with motion and appearance.
The overall methodology is illustrated in Fig.~\ref{fig:overview}.
Our approach first predicts temporally consistent masks and generate instance captions with MLLMs, using both as supervision during training to enforce identity consistency and semantic alignment~(Sec.~\ref{sec:31}).
We then explicitly disentangle foreground and background components.
For the static background, we aggregate observations across the entire temporal sequence and jointly optimize a temporally consistent Gaussian representation $\mathcal{G}_{bg}$ to ensure stable spatio-temporal reconstruction.
For the foreground, we equip each Gaussian with a learnable semantic feature and use two MLP heads to predict instance identity and language-aligned embedding~(Sec.~\ref{sec:32}). Guided by optical flow, we then learn a unified dynamic representation $\mathcal{G}_{fg}$ that jointly models motion (position $p$, rotation $q$), appearance (third-order spherical harmonic $\mathcal{C}$, opacity $\sigma$, and scaling $s$), and semantic feature $f$ attributes in an end-to-end manner. 
This unified 4D semantic formulation, optimized with SDF constraints and regularization terms, ensures temporal coherence and photorealistic rendering, while laying a solid foundation for instance-aware editing and open-vocabulary querying~(Sec.~\ref{sec:33}).

\subsection{Multimodal Instance-level Initialization} \label{sec:31}
Existing 4D representation methods~\cite{jiang2024hifi4g, dualgs, taogs} heavily rely on background matting~\cite{BGMv2}, which is unreliable in sports scenarios due to dynamic lighting and reflections.
To mitigate this issue, we employ SAM3~\cite{sam3} and use text prompts such as “basketball player” and “basketball” to extract instances masks for basketball scenes.
Leveraging the memory bank mechanism of SAM3, we obtain temporally consistent instance identities within each video.
To further establish cross-view identity consistency, we concatenate the first frame from each view into a canonical video and reapply SAM3. Since adjacent cameras capture highly overlapping content from closely spaced viewpoints, the resulting canonical masks are naturally aligned and provide a unified identity reference.
We compute the IoU between canonical and per-view first-frame masks to establish one-to-one correspondence, and propagate the aligned identities to all frames, yielding temporally and spatially consistent instance masks across all views. 
Formally, given $V$ views, $T$ frames and $D$ instances, we denote the segmentation of view $v$ at frame $t$ as a multi-class mask $M^{v, t} \in\{0,1, \ldots, D\}$. The background region is defined as $M_{b g}^{v, t} = M^{v, t}_0$, while $M^{v, t}_d, d \in\{1, \ldots, D\}$ corresponds to the d-th instance identity.

While instance-level geometric partitioning provides spatially grounded identity, it lacks high-level semantic abstractions aligned with natural language.
To bridge this gap, we generate instance-wise captions and encode them into embeddings, providing language-aligned supervision for the dynamic semantic field.

Recent MLLMs, such as GPT-4o~\cite{hurst2024gpt}, Qwen3-VL~\cite{bai2025qwen3}, enable high-quality language generation from multimodal inputs. 
Leveraging the capabilities, we combine visual cues and textual prompts to guide the MLLM in producing temporally consistent and object-specific captions across video frames, capturing appearance, position, and motion patterns.
Inspired by 4DLangsplat~\cite{li20254d}, we generate both time-invariant global captions and time-dependent per-frame captions for each instance, followed by language embedding extraction.
For the time-invariant global caption $C_i^{\text{global}}$, we highlight the instance in the first view using the SAM3 mask rendered as a red contour and prompt Qwen3-VL~\cite{bai2025qwen3} to describe its overall appearance and action characteristics.
For the per-frame caption, we aggregate all views at frame $t$ into a spatially consistent video clip and condition the MLLM on the highlighted instance observations $\bar{\mathcal{P}}_i^{v,t}$, the global caption, and a temporal prompt $\mathcal{T}$:
\begin{equation}
C_i^{t} 
= \text{MLLM}\big( \{\bar{\mathcal{P}}_i^{v,t}  \}_{v=0}^{V-1},\; C_i^{\text{global}},\; \mathcal{T} \big).
\end{equation}
The generated time-dependent captions are encoded into high-dimensional embeddings using an LLM~\cite{wang2024improving} and further compressed into a compact 6-dimensional embeddings via a lightweight autoencoder.
The embeddings $e$, together with the instance masks, serve as 2D semantic supervision for learning the dynamic semantic field, enforcing temporally consistent representations across frames.

\subsection{Gaussian Initialization} \label{sec:32}
With the aligned instance masks and semantic embeddings as guidance, we explicitly decompose the scene into static background and dynamic foreground, and optimize each layer, $\mathcal{G}_{bg}$ and $\mathcal{G}_{fg}$, with tailored training strategies to enhance consistency and achieve detailed 4D segmentation.

\noindent{\bf Static Background Initialization.}
Our goal is to explicitly model the temporally consistent components of the scene, facilitating more accurate foreground modeling and separation. To this end, we uniformly sample ground-truth images across both temporal and spatial dimensions, and jointly optimize global background Gaussians $\mathcal{G}_{bg}$ using the following loss:
\begin{equation}
\begin{aligned}   
E^{bg}_{\mathrm {color }}&=\|M_{bg}(\hat{\mathbf{C}}-\mathbf{C})\|_1, \\
E_{\mathrm{bg}}&= \lambda_{\mathrm{iso}} E_{\mathrm{iso}}+\lambda_{\mathrm{size}} E_{\mathrm{size}}+E^{bg}_{\mathrm{color}}, 
\end{aligned}
\end{equation}
where the color loss $E^{bg}_{\mathrm {color }}$ is evaluated exclusively over the background mask $M_{bg}$. $\hat{\mathbf{C}}$ denotes the blended color after rasterization, and $\mathbf{C}$ is the ground truth.
The isotropic loss $E_{\mathrm{iso}}$ and the size loss $E_{\mathrm{size}}$ regularize the scaling of the Gaussians, preventing them from becoming overly elongated or excessively large. We refer to DualGS~\cite{dualgs} for the detailed formulation. The background layer provides a global reference for subsequent optimization, while the Gaussians in the background are also continuously trained in each frame.

\noindent{\bf First Frame Initialization.} 
Our objective is to construct a temporally coherent 4D representation that jointly models dynamic motion and photometric appearance, while preserving instance-level structure and higher-level semantic embeddings.
To achieve this, we equip the Gaussian representation with learnable semantic feature attributes $f$ that explicitly encode instance-level semantic information. 
Each Gaussian primitive is associated with a learnable, view-independent semantic attribute vector $f$ (8-dimensional in our experiments), initialized by randomly sampling from a uniform distribution. 

During optimization, these semantic features are jointly optimized alongside other Gaussian parameters to encode the instance identities in the scene. 
Given the pre-trained background layer, we optimize $\mathcal{G}_{bg}$ and $\mathcal{G}_{fg}$ together by feeding both into the differentiable Gaussian rasterization. In addition to rendering the color map via alpha blending, we also rasterize the semantic attributes to produce a 2D semantic feature map $F \in \mathbb{R}^{H \times W \times 8}$. 
To supervise instance identity, $F$ is projected through a classification MLP decoder into $(D+1)$ channels, followed by a softmax to produce per-pixel class probabilities ${P}$. These per-pixel probabilities are supervised with instance masks $M$ via a cross-entropy loss:
\begin{equation}
\begin{aligned}   
E_{\text {id }}&=-\sum_{d=0}^D y_d \log {P}_d, 
\end{aligned}
\end{equation}
where $y_d$ is the one-hot encoding of the ground-truth instance label.
We also utilize a separate MLP decoder to produce language-aligned embeddings $\hat e$, which are supervised with the ground-truth embeddings $e$ via an L1 loss:
\begin{equation}
\begin{aligned}   
E_{\text {emb }}&=\|\hat e -e\|_1, 
\end{aligned}
\end{equation}
This dual-decoder design ensures that the model simultaneously captures low-level instance identities and high-level semantic information.

To further enforce the spatial consistency of the semantic feature, we follow the Gaussian Grouping~\cite{GaussianGrouping} to introduce the 3D regularization term:
\begin{equation}
\begin{aligned}   
{E}_{3 \mathrm{D}}=\frac{1}{|\mathcal{S}|} \sum_{i \in \mathcal{S}} \frac{1}{k} \sum_{j \in \mathcal{N}_k(i)} D_{\mathrm{KL}}\left(P_i \| P_j\right),
\end{aligned}
\end{equation}
where $\mathcal{S}$ denotes the sampled foreground Gaussians, $\mathcal{N}_k(i)$ is the set of $k$-nearest Gaussians of $i$ in 3D space, and $D_{\mathrm{KL}}(\cdot \| \cdot)$ is the Kullback-Leibler divergence. This loss enforces spatial smoothness by encouraging neighboring Gaussians within the same instance to learn consistent identity representations, enabling occluded or interior regions to acquire coherent features that facilitate flexible scene manipulation.
Overall, the objective function in the first frame training is formulated as follows:
\begin{equation}
\begin{aligned}   
E_{\mathrm{init}}= \lambda_{\mathrm{id}} E_{\mathrm{id}}+ \lambda_{\mathrm{emb}} E_{\mathrm{emb}} + \lambda_{\mathrm{3D}} E_{\mathrm{3D}}+E_{\mathrm{color}},
\end{aligned}
\end{equation}
where the color loss $E_{color}$ follows the original 3DGS~\cite{3dgs}.

\subsection{Dynamic Optimization} \label{sec:33}

After initialization, we adopt a frame-by-frame optimization strategy that unifies tracking, modeling, and semantic learning within a single representation and trains them jointly. The optimization process consists of two stages: optical-flow-aided explicit warping, and the training phase.

\noindent{\bf Explicit Warping.} Assuming that the foreground Gaussians $\mathcal{G}_{fg}^{t-1}$ from the preceding frame have been well-optimized, we aim to explicitly warp each primitive with position $p_{t-1}$ to $\bar{p_t}$, providing a reliable initialization for the current frame $t$. Since optical flow operates in the 2D image space, we establish geometric constraints through projection and multi-view least squares.
Specifically, we project $p_{t-1}$ onto each camera plane to obtain pixel coordinates $u=(x_v,y_v)$, for $v=0,...,V-1$. These 2D coordinates are then used to query the optical flow predicted by SEA-RAFT~\cite{wang2024sea} between frame $t-1$ and $t$, yielding the warped pixel locations $u_v' = (x'_v,y'_v)$. These pixel positions indicate the potential projections of the Gaussian position in the current frame. We therefore initialize the updated 3D position $\bar{p_t}$ by minimizing the multi-view reprojection error:
\begin{equation}
    \bar{p}_t = \arg\min_{p \in \mathbb{R}^3} \sum_{v=0}^{V-1} \left\| \pi_v(p) - u'_v \right\|_2^2,
\end{equation}
where $\pi_v(\cdot)$ denotes the perspective projection under view $v$. This least-squares problem can be further converted into solving a linear equation.

\noindent{\bf Training.} 
Notably, the least-squares triangulation is inaccurate due to noise and occlusions, which can introduce incorrect correspondences and lead to implausible initialization. Inspired by HiFi4G~\cite{jiang2024hifi4g}, we introduce a refinement stage that employs a locally as-rigid-as-possible (ARAP) regularizer to enforce local geometric consistency:
\begin{equation}
\begin{aligned}
E_{\text{smooth}} = \sum_{i} \sum_{k \in \mathcal{N}(i)} w_{i,k}^{t-1} \| & R\left(q_{i,t} * q_{i,t-1}^{-1}\right) \left(p_{k,t-1} - p_{i,t-1}\right) - \left(p_{k,t} - p_{i,t}\right) \|_{2}^{2},
\end{aligned}
\end{equation}
where $i$ indexes a Gaussian primitive and $R(\cdot)$ converts a quaternion to a rotation matrix and $w$ corresponds to the blending weights $w_{i,k}^{t} = \exp \left(-\left\|p_{i,t} - p_{k,t}\right\|_{2}^{2} / l^{2}\right)$. $l$ is the influence radius.
In this phase, we keep the appearance and semantic attributes fixed and optimize only the Gaussian positions and rotations to correct physically implausible deformations caused by explicit warping. The objective is formulated as:
\begin{equation}
\begin{aligned}
E_{\mathrm{ft}}= \lambda_{\mathrm{smooth}} E_{\mathrm{smooth}}+E_{\mathrm{color}},
\end{aligned}
\end{equation}

After refinement, we jointly optimize the Gaussians’ motion, appearance, and semantic feature attributes. 
Foreground Gaussian primitives dynamically split and prune to accommodate new observations, while cloned Gaussians inherit all parents attributes, ensuring continuous trajectory tracking. 
To maintain semantic feature consistency, we constrain each Gaussian’s 2D projections to remain within its assigned instance masks, promoting effective learning.
For each instance $d$, we first pre-compute a signed distance field (SDF) $\Phi^{v,t}_{d}$ from the its instance mask $M^{v,t}_{d}$, where each pixel $u$ is assigned the Euclidean distance to the nearest mask boundary. 
Recall that the classification MLP produces per-pixel semantic logits, which can also be applied to each Gaussian’s semantic feature. We then compute a softmax to predict Gaussian
 $i$'s instance label $c_i$, and define the SDF loss only for Gaussians whose predicted label matches instance $d$, defined as:
\begin{equation}
    E_{\text{sdf}} = \sum_{i} y_d \cdot \left[ \max(0, \Phi^{v,t}_{d}(\pi_v(p_{i,t}))) \right]^2,
\end{equation}
This formulation penalizes Gaussians that lie outside the 2D silhouette of their assigned instance, reinforcing spatial consistency.
Besides, we also apply several regularization terms to maintain temporal consistency. For the background layer, we continuously optimize the appearance attributes $a_{i,t}$ to capture potential lighting and shadow changes, while constraining them to remain close to the global background appearance attributes $a_{i}$ from $\mathcal{G}_{bg}$ to prevent jitter:
\begin{equation}
    E_{\text{temp}}^{bg} = \sum_{i} \left\| a_{i,t} - a_{i} \right\|_2^2,
\end{equation}
We further extend this regularization across consecutive frames, denoted as $E_{\text{temp}}$, to prevent abrupt changes in all attributes, avoiding unstable tracking or sudden semantic shifts.
Overall, the training-phase loss is defined as:
\begin{equation}
\begin{aligned}
E_{\mathrm{train}}=& \lambda_{\mathrm{sdf}} E_{\mathrm{sdf}}+ \lambda_{\text{temp}}^{bg} E_{\text{temp}}^{bg} + \lambda_{\text{temp}} E_{\text{temp}}  + \lambda_{\mathrm{smooth}} E_{\mathrm{smooth}}+ \\&\lambda_{\mathrm{id}} E_{\mathrm{id}} + \lambda_{\mathrm{emb}} E_{\mathrm{emb}} +\lambda_{\mathrm{3D}} E_{\mathrm{3D}}+E_{\mathrm{color}}.
\end{aligned}
\end{equation}

\noindent{\bf Querying.} 
After training, we decode the rendered feature map into high-dimensional space and compute relevance scores via cosine similarity with the LLM-processed query. Specifically, we first retrieve the instance ID in the first frame using the identity query, then render the RGB and feature maps for that instance only, and finally retrieve its most relevant frame segments via the segment query. This approach effectively filters out interference from other dynamic instances, enabling precise instance-level retrieval and analysis.

\begin{figure*}[t] 
	\begin{center} 
        \includegraphics[width=0.95\linewidth]{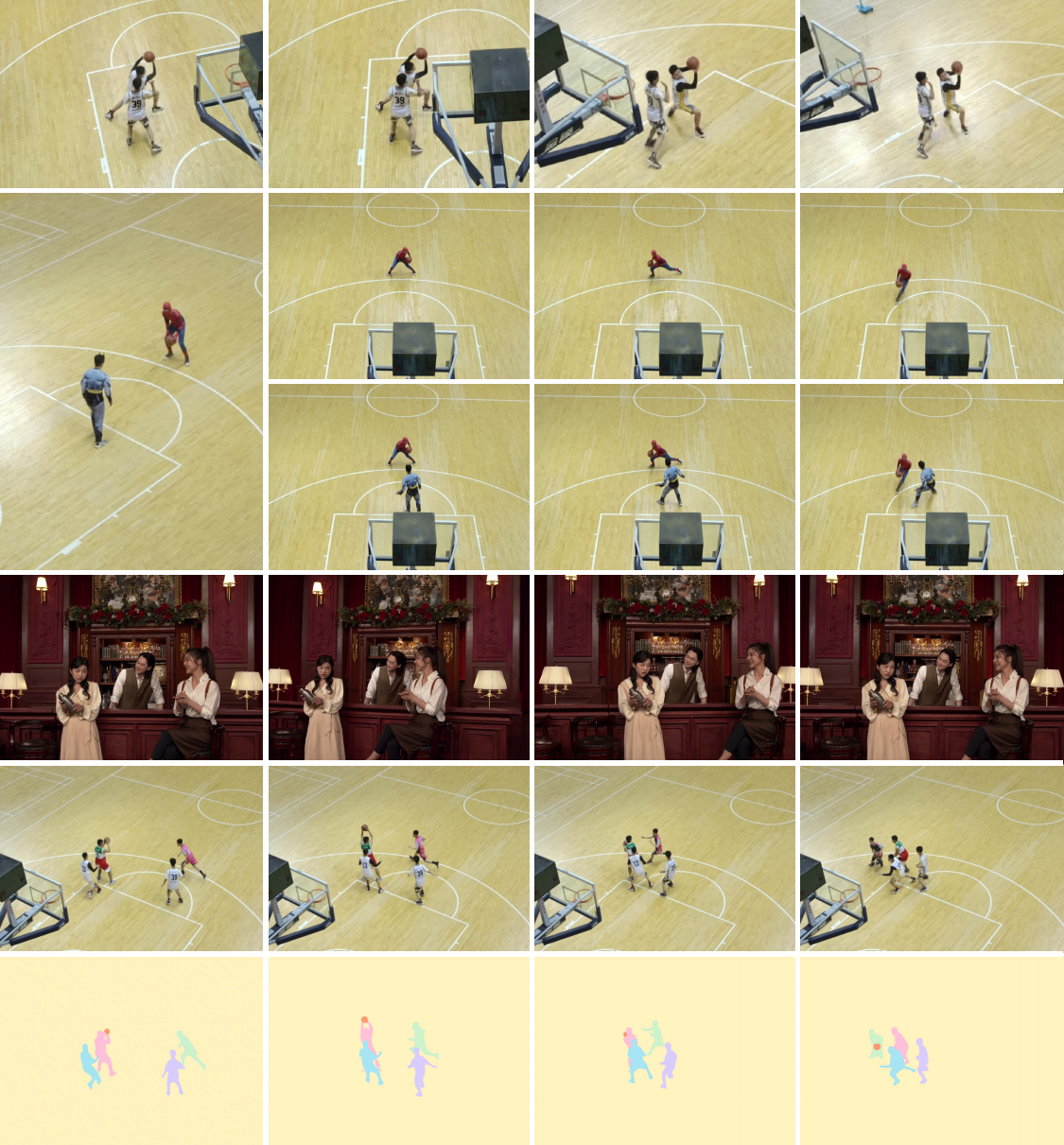} 
	\end{center} 
	\vspace{-15pt}
    \caption{Gallery of our results. All images are rendered from novel views. The first four rows show high-fidelity rendering results under challenging scenarios, including fast motions and severe occlusions. The fifth row presents the corresponding instance-level 4D segmentation of the fourth row.}
	\label{fig:gallery}
	\vspace{-25pt}
\end{figure*}

\noindent{\bf Implementation Details.}  
For static background initialization, we set $\lambda_{\mathrm{iso}} = 0.0005$ and $\lambda_{\mathrm{size}} = 0.02$, optimizing for 20,000 iterations.  
For first frame initialization, we jointly optimize the Gaussian and two MLPs for 30,000 iterations, using $\lambda_{\mathrm{id}} = 2$, $\lambda_{\mathrm{emb}} = 10$, and $\lambda_{\mathrm{3D}} = 2$. For the $E_{3D}$, we sample $6000$ foreground Gaussians and use $k=4$ nearest neighbors for KL divergence. 
During the refinement stage, we apply $\lambda_{\mathrm{smooth}} = 0.01$ and finetune for 3000 iterations. 
In the dynamic training stage, we adopt the following empirical weights: $\lambda_{\mathrm{sdf}} = 0.01$, $\lambda_{\text{temp}}^{bg} = 0.001$, $\lambda_{\text{temp}} = 0.01$, $\lambda_{\mathrm{smooth}} = 0.0001$, $\lambda_{\mathrm{id}} = 1$, $\lambda_{\mathrm{emb}} = 10$, and $\lambda_{\mathrm{3D}} = 2$. We train each frame for 8000 iterations. During the training, we freeze the classification MLP, continue optimizing only the semantic MLP. Every 300 iterations, we clone and prune foreground Gaussians, ensuring that the total number of modified Gaussians per frame remains below $5\%$.

\section{Experimental Results} 

To demonstrate the capabilities of \abbtitle, we evaluate our method on two challenging datasets: the ST-NeRF basketball dataset~\cite{zhang2021editable} and the MPEG GSC Dataset~\cite{jeong2024invr}. The basketball dataset is captured using 32 pre-synchronized and calibrated Z-Cam cameras, recording 4K videos at 30 fps. It features multi-player basketball games with fast and complex motions, as well as challenging multi-person interactions such as screens, defensive switches, fast-break layups, and jump shots. The MPEG GSC Dataset~\cite{jeong2024invr} contains two indoor multi-view sequences featuring fast motion scenarios, captured by 20 cameras at 1080p resolution.
We conduct all experiments on a single NVIDIA RTX 3090 GPU. Background and first-frame initialization take about 50 minutes, while dynamic optimization takes roughly 10 minutes per frame.
As illustrated in Fig.~\ref{fig:gallery}, our method enables robust tracking and high-fidelity rendering, while also producing temporally consistent 4D instance segmentation for scene editing. This allows flexible manipulation of dynamic elements, such as changing an instance’s position, size, or orientation, or removing it entirely.

\subsection{Comparison} 
\begin{figure*}[t] 
	\begin{center} 
		\includegraphics[width=0.99\linewidth]{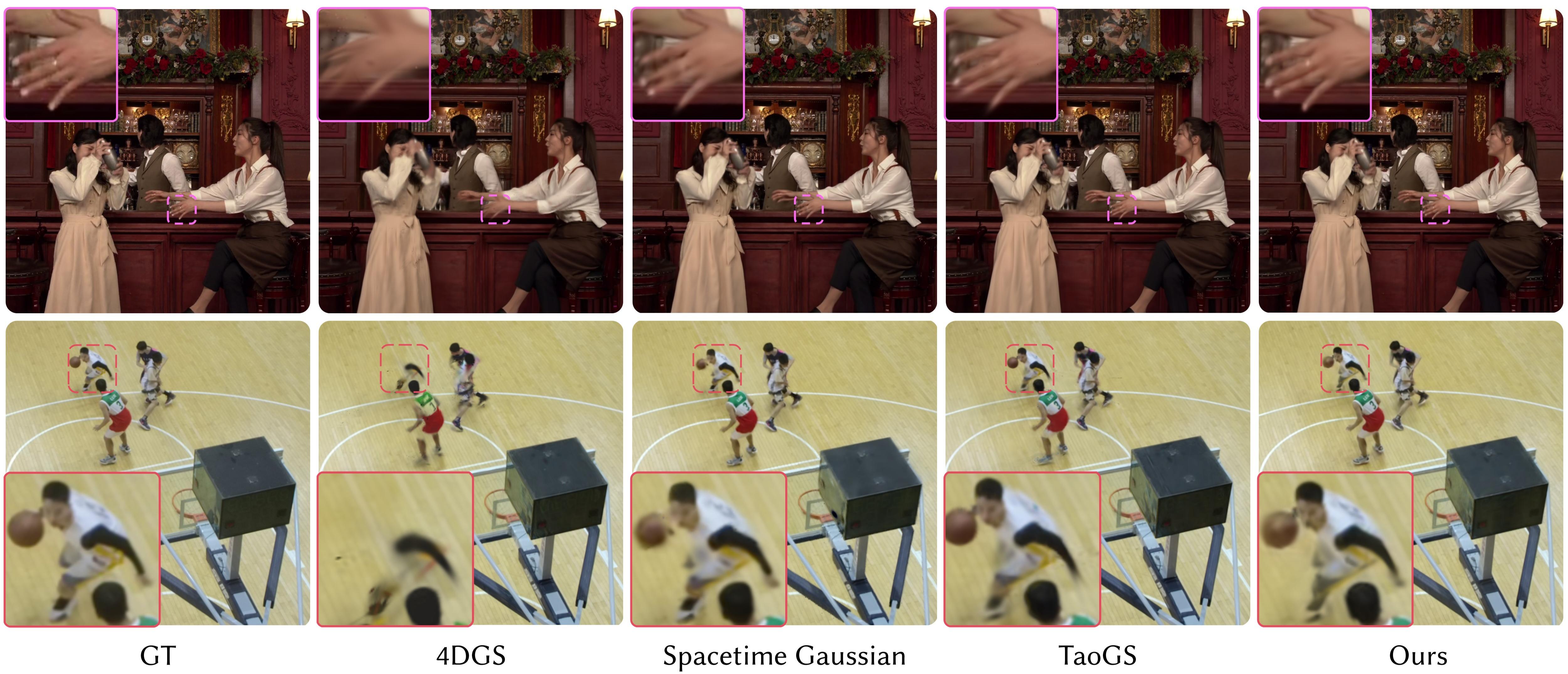} 
	\end{center} 
	\vspace{-16pt}
	\caption{Qualitative comparison with 4DGS~\cite{Wu_2024_CVPR}, Spacetime Gaussian~\cite{li2023spacetime}, and TaoGS~\cite{taogs}. Ours shows the best quality. Note that the zoomed-in regions are extremely small in the original image, where even the ground truth is not very sharp.}
	\label{fig:fig_comp_1}
    \vspace{-10pt}

\end{figure*} 

\begin{figure*}[t] 
	\begin{center} 
		\includegraphics[width=0.99\linewidth]{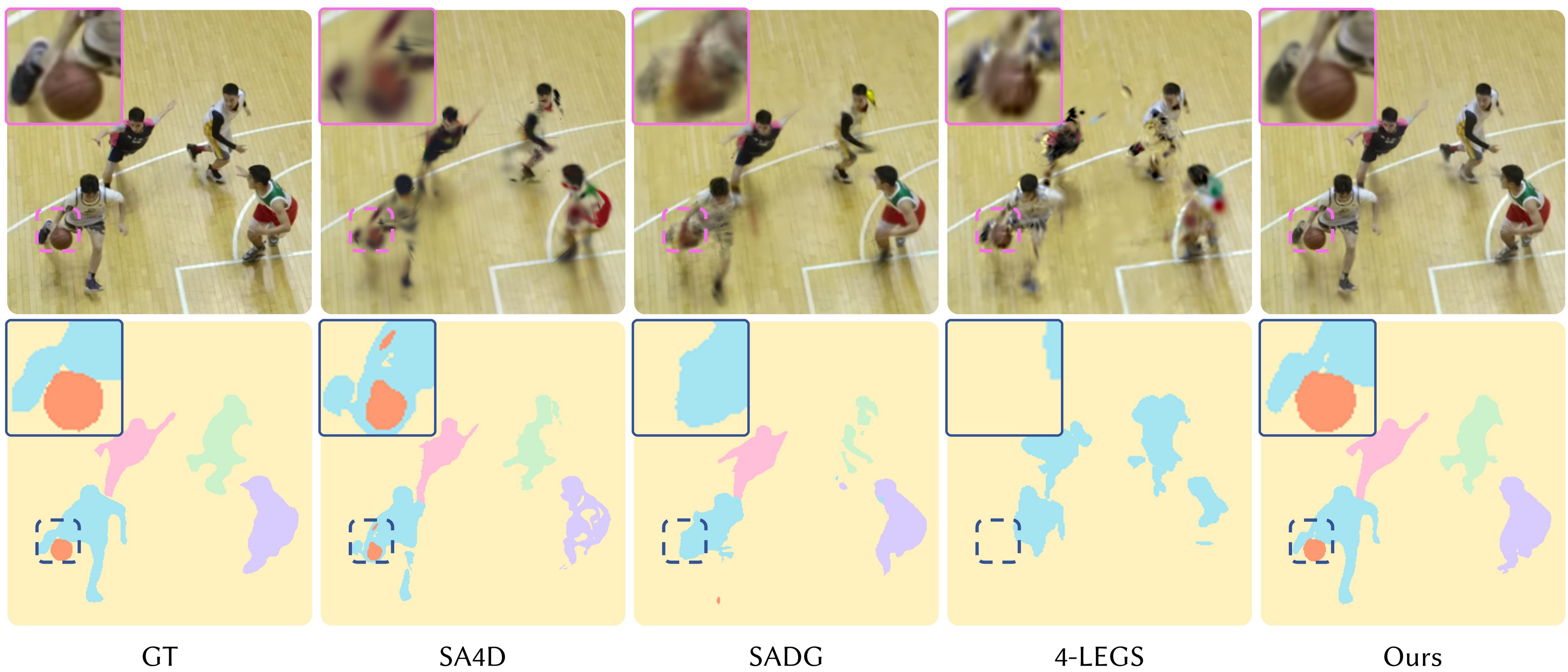} 
	\end{center} 
	\vspace{-16pt}
	\caption{Qualitative comparison with 4D segmentation methods, including SA4D~\cite{sa4d}, SADG~\cite{li2024sadg}, and 4-LEGS~\cite{fiebelman20254legs}.Ours achieves more accurate instance segmentation.}
	\label{fig:fig_comp_2}
    \vspace{-11pt}

\end{figure*} 

\begin{table}[t]
    \centering
    \caption{Rendering quality comparison on the corresponding dataset. Green and yellow cells indicate the best and second-best results, respectively.}
    \resizebox{0.90\textwidth}{!}{
    \begin{tabular}{l ccc ccc}
        \toprule
        \multirow{2}{*}{Method} & \multicolumn{3}{c}{ST-NeRF} & \multicolumn{3}{c}{MPEG GSC} \\
        \cmidrule(lr){2-4} \cmidrule(lr){5-7}
        & PSNR $\uparrow$ & SSIM $\uparrow$ & LPIPS $\downarrow$ 
        & PSNR $\uparrow$ & SSIM $\uparrow$ & LPIPS $\downarrow$ \\
        \midrule
        4DGS 
        & 34.722 & 0.956 & 0.0625 
        & 33.542 & 0.936 & 0.0912 \\

        Spacetime Gaussian 
        & \secondBestCellColor{37.949} & \secondBestCellColor{0.965} & \secondBestCellColor{0.0524} 
        & \secondBestCellColor{37.807} & \secondBestCellColor{0.954} & \bestCellColor{0.0526} \\

        TaoGS
        & 37.719 & 0.963 & 0.0564 
        & 36.137 & 0.952 & 0.0671 \\

        \midrule
        Ours 
        & \bestCellColor{38.912} & \bestCellColor{0.967} & \bestCellColor{0.0463} 
        & \bestCellColor{38.241} & \bestCellColor{0.959} & \secondBestCellColor{0.0537} \\
        \bottomrule
    \end{tabular}
    }
        \vspace{-23pt}

    \label{table:comparison_render}
\end{table}

\begin{table}[t]
    \centering
    \caption{Rendering quality and segmentation comparison on the basketball dataset.}
    \label{table:comparison_mask}
    
    \small %
    \setlength{\tabcolsep}{5pt} %
    
    \begin{tabular}{l ccc ccc}
        \hline
        Method & PSNR $\uparrow$ & SSIM $\uparrow$ & LPIPS $\downarrow$ & mIoU $\uparrow$ & Recall $\uparrow$ & $F_1$ $\uparrow$ \\
        \hline
        SA4D & \secondBestCellColor{33.951} & \secondBestCellColor{0.942} & 0.0917 & \secondBestCellColor{0.5235} & \secondBestCellColor{0.5667} & \secondBestCellColor{0.6464} \\
        SADG & 33.421 & \secondBestCellColor{0.942} & 0.1259 & 0.4995 & 0.5330 & 0.6052\\
        4-LEGS  & 31.655 & 0.939 & \secondBestCellColor{0.0784} & -- & -- & --\\
        \hline
        Ours & \bestCellColor{38.298} & \bestCellColor{0.965} & \bestCellColor{0.0470} & \bestCellColor{0.8305} & \bestCellColor{0.8764} & \bestCellColor{0.8878}  \\
        \hline
    \end{tabular}
    \vspace{-10pt} %
\end{table}

\begin{figure}[t] 
	\vspace{5pt}

	\begin{center} 
		\includegraphics[width=1\linewidth]{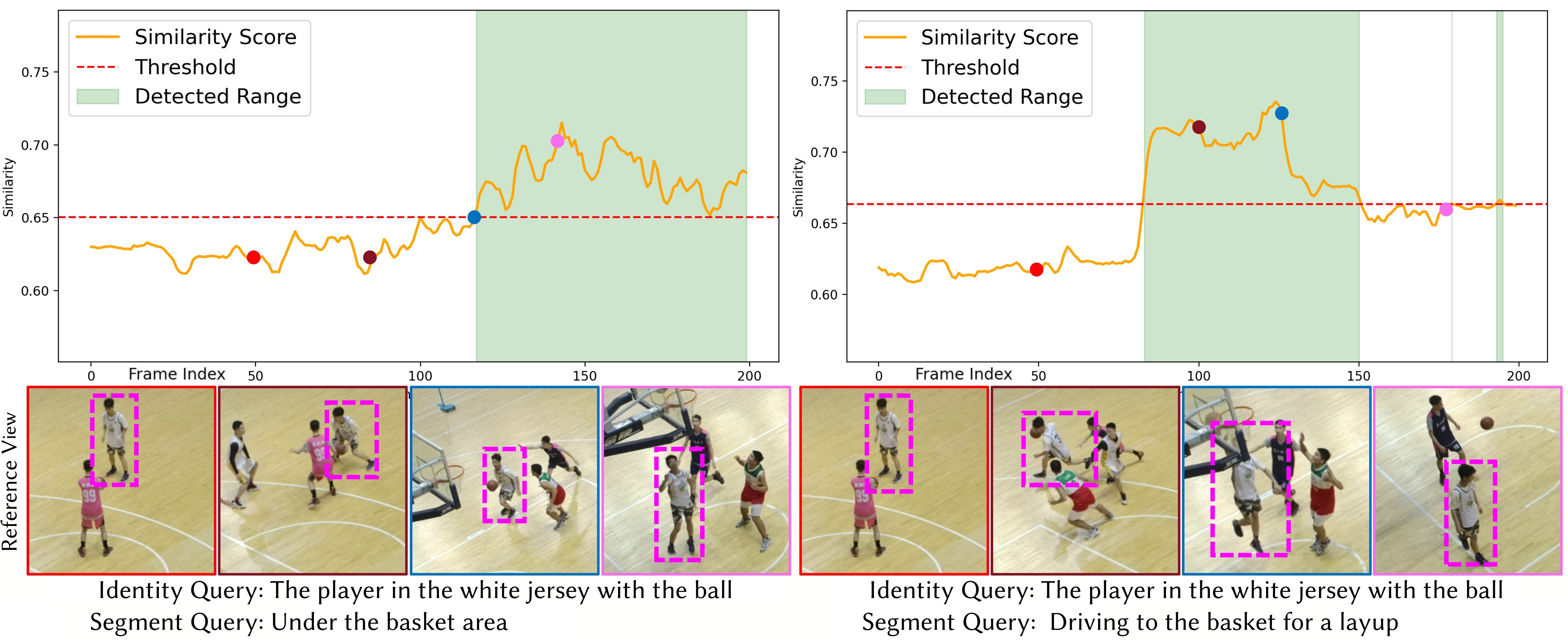}
	\end{center} 
    \vspace{-15pt}
	\caption{Visualization of querying results. The top row shows the query–frame similarity scores across frames. The bottom row presents several reference views.} 
	\label{fig:vis}
	\vspace{-20pt}
\end{figure}

\begin{figure}[t] 
	\vspace{10pt}

	\begin{center} 
		\includegraphics[width=1\linewidth]{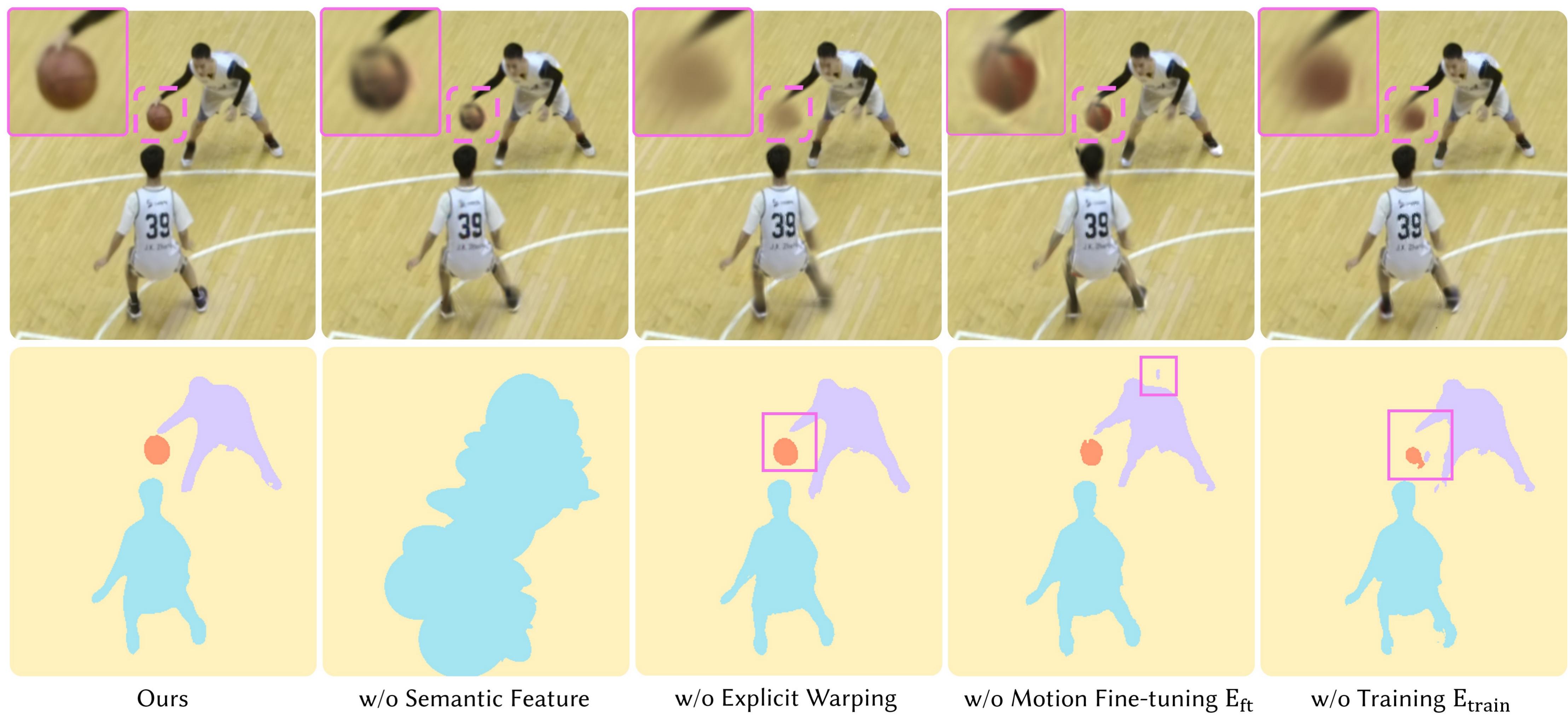}
	\end{center} 
    \vspace{-14pt}
	\caption{Qualitative ablation study on key training components.} 
	\label{fig:ab_train}
	\vspace{-15pt}
\end{figure}

\begin{table}[t]
    \centering
    \vspace{0.1in}
    
    \begin{minipage}[t]{0.48\linewidth}
        \centering
        \caption{Quantitative ablation study on training stages.}
        \label{table:ab_train}
        \vspace{5pt}
        \setlength{\tabcolsep}{4pt} %
        \resizebox{\textwidth}{!}{
            \begin{tabular}{l|cccc}
                \hline
                & PSNR $\uparrow$ & mIoU $\uparrow$ & Recall $\uparrow$ & $F_1$ $\uparrow$ \\
                \hline
                w/o Semantic Feature & \secondBestCellColor{37.661} & -- & -- & -- \\
                w/o Explicit Warping & 37.091 & \secondBestCellColor{0.8008} & \secondBestCellColor{0.8420} & \secondBestCellColor{0.8743} \\
                w/o Motion Fine-tuning $E_{\mathrm{ft}}$ & 37.618 & 0.6987 & 0.7228 & 0.7881 \\
                w/o Training $E_{\mathrm{train}}$  & 36.796 & 0.5105 & 0.5233 & 0.6065 \\
                \hline
                Ours & \bestCellColor{\textbf{38.382}} & \bestCellColor{\textbf{0.8328}} & \bestCellColor{\textbf{0.8690}} & \bestCellColor{\textbf{0.8940}} \\
                \hline
            \end{tabular}
        }
    \end{minipage}
    \hfill
    \begin{minipage}[t]{0.48\linewidth}
        \centering
        \caption{Quantitative comparison of different loss term contributions.}
        \label{table:ab_term}
        \vspace{5pt}
        \setlength{\tabcolsep}{4pt}
        \resizebox{\textwidth}{!}{
            \begin{tabular}{l|cccc}
                \hline
                & PSNR $\uparrow$ & mIoU $\uparrow$ & Recall $\uparrow$ & $F_1$ $\uparrow$ \\
                \hline
                w/o $E_{\mathrm{sdf}}$ & 37.604 & 0.7404 & 0.7780 & 0.7952 \\
                w/o $E_{\text{temp}}^{bg}$ & \secondBestCellColor{37.967} & \secondBestCellColor{0.8095} & 0.8548 & \secondBestCellColor{0.8693} \\
                w/o $E_{\text{temp}}$ & 37.529 & 0.7613 & 0.8284 & 0.8370 \\
                w/o $E_{\mathrm{smooth}}$ &37.216 & 0.7915 & \secondBestCellColor{0.8653} & 0.8631  \\
                \hline
                Ours & \bestCellColor{\textbf{38.284}} & \bestCellColor{\textbf{0.8341}} & \bestCellColor{\textbf{0.8794}} & \bestCellColor{\textbf{0.8907}} \\
                \hline
            \end{tabular}
        }
    \end{minipage}
    \vspace{-10pt}
\end{table}

\myparagraph{Qualitative Evaluation.}
We compare \abbtitle against state-of-the-art dynamic rendering methods, including 4DGS~\cite{Wu_2024_CVPR}, Spacetime Gaussian~\cite{li2023spacetime}, and TaoGS~\cite{taogs}. As shown in Fig.~\ref{fig:fig_comp_1}, 4DGS struggles to model fast motions effectively, while Spacetime Gaussian produces blurry results. TaoGS suffers from over-smoothing, leading to noticeable artifacts. In contrast, \abbtitle explicitly decomposes the scene, leverages optical-flow guidance, and incorporates regularization terms to achieve high-fidelity rendering even under rapid and challenging motions.
To further validate the 4D segmentation performance, we compare our method against with three baselines, SA4D~\cite{sa4d}, SADG~\cite{li2024sadg} and 4-LEGS~\cite{fiebelman20254legs}, all provided with SAM3 masks as input. As shown in Fig.~\ref{fig:fig_comp_2}, SA4D struggles with fast motions, resulting in blurred and incomplete masks. SADG also has difficulty with rapid movements and often merges closely interacting instances. 4-LEGS produces inferior rendering quality and supports only foreground-level mask queries, with instance-level queries fail entirely. In contrast, our method delivers high-quality renderings while producing 4D segmentation results that are visually more accurate than the SAM3 inputs.

\myparagraph{Quantitative Comparison.}
We select three 200-frame sequences from the ST-NeRF basketball dataset and two 100-frame sequences from the MPEG GSC dataset, evaluating rendering performance across three test views.
For rendering, we evaluate PSNR, SSIM, LPIPS in Tab.~\ref{table:comparison_render}. Our method achieves the highest rendering quality, surpassing existing approaches.
To assess 4D segmentation, we evaluate mIoU, Recall, and F1 on instance masks, where mIoU measures mask overlap, Recall quantifies correctly predicted instance pixels, and F1 balances precision and recall. Instance-level masks for 4-LEGS are difficult to obtain, so we exclude it from this comparison. As shown in Tab.~\ref{table:comparison_mask}, our method achieves the highest 4D segmentation accuracy while maintaining superior rendering quality.

We also visualize the temporal evolution of query-frame similarity scores in Fig.~\ref{fig:vis}, using the average across all frames as the threshold. For reference, we also provide the raw image with the target instance highlighted.

\subsection{Ablation Study} \label{sec:abla} 

\myparagraph{Key Components.} 
We perform an ablation study to evaluate the impact of key training components in our optimization. As shown in Fig.~\ref{fig:ab_train} and Tab.~\ref{table:ab_train}, disabling semantic features leads to under-constrained 4D Gaussian shapes, blurred renderings, and failed instance-wise segmentation. Omitting Optical-Flow-Aided Explicit Warping causes failures in regions with fast motion. Skipping motion fine-tuning $E_{\mathrm{ft}}$ leads to artifacts that disrupt temporal tracking, while removing the training $E_{\mathrm{train}}$ leads to severe blurring. In contrast, our full method effectively leverages both RGB and semantic supervision, producing clear and temporally consistent reconstructions.

\myparagraph{Regularizers.} We evaluate the influence of regularizers during training. As shown in Tab.~\ref{table:ab_term}, the SDF term constrains Gaussians to lie within the instance silhouette, the temporal term enforces consistency across frames, and the smoothness term ensures physically plausible non-rigid deformations. Omitting any of these terms degrades performance. In contrast, our full method produces high-quality rendering results and accurate 4D segmentation.

\section{Limitations}
Despite \abbtitle enabling high-fidelity rendering, robust tracking, and accurate segmentation for dynamic scenes, it has several limitations.
First, training dynamic Gaussians remains relatively slow, which limits practical applications. Accelerating this step is an important direction for future work.
Second, our framework relies on multiple loss terms with carefully tuned hyperparameters. Adapting these to significantly different scenes may require additional tuning.
Finally, due to current computational constraints, we encode semantic features into a compact, low-dimensional latent space to enable Gaussian training. This limits the richness of semantic information that can be represented.

\section {Conclusion}

In this work, we presented \abbtitle, a unified 4D Gaussian framework that tightly integrates geometric modeling and semantic understanding for highly dynamic scenes. 
By leveraging spatially and temporally consistent instance masks alongside language-aligned embeddings from MLLMs, our method achieves robust human performance tracking, high-fidelity rendering, and precise instance-level segmentation. 
The proposed two-layer Gaussian representation, combined with an optical-flow-aided optimization strategy, ensures high temporal consistency even under fast and complex motions. 
Experiments demonstrate that \abbtitle effectively couples reconstruction and semantic reasoning, enabling open-vocabulary queries and representing a significant step toward more immersive virtual experiences.

\bibliographystyle{splncs04}
\bibliography{main}

\end{document}